\documentclass[journal]{IEEEtran}

\usepackage{comment}
\usepackage{epsfig}
\usepackage{graphicx}
\usepackage{amsmath}
\usepackage{amssymb}
\usepackage{comment}
\usepackage{multirow}

\usepackage{hyperref}
\hypersetup{hidelinks}

\usepackage{booktabs}
\usepackage{array, caption, threeparttable}
\usepackage{caption}
\usepackage{subfigure}
\usepackage{algorithm}
\usepackage{listings}
\usepackage{color}
\usepackage{mathtools}

\usepackage{todonotes}
\usepackage{gensymb}

\def\etal{\emph{et al}.}

\begin{document}

\title{Searching Multi-Rate and Multi-Modal Temporal Enhanced Networks for Gesture Recognition}

\author{Zitong Yu, Benjia Zhou, Jun Wan,~\IEEEmembership{Senior Member,~IEEE}, Pichao Wang, Haoyu Chen, Xin Liu, \\
Stan Z. Li,~\IEEEmembership{Fellow,~IEEE} and Guoying Zhao,~\IEEEmembership{Senior Member,~IEEE} 

\thanks{The first two authors contribute equally. Corresponding authors: Jun Wan \{jun.wan\}@nlpr.ia.ac.cn, and Guoying Zhao \{guoying.zhao\}@oulu.fi.}

\thanks{Z. Yu, H. Chen, X. Liu and G. Zhao are with Center for Machine Vision and Signal Analysis, University of Oulu, Oulu 90014, Finland. }


\thanks{B. Zhou is with Macau University of Science and Technology, Macau 999078, China.}
\thanks{J. Wan is with National Laboratory of Pattern Recognition, Institute of Automation, Chinese Academy of Sciences, Beijing 100190, China. }
\thanks{P. Wang is with DAMO Academy, Alibaba Group (U.S.) Inc., Bellevue, WA, 98004, USA.}

\thanks{Stan Z. Li is with Westlake University, Hangzhou 310012, China. }}

\markboth{Journal of \LaTeX\ Class Files,~Vol.~14, No.~8, August~2015}%
{Shell \MakeLowercase{\textit{et al.}}: Bare Demo of IEEEtran.cls for IEEE Journals}

\maketitle

\begin{abstract}
Gesture recognition has attracted considerable attention owing to its great potential in applications. Although the great progress has been made recently in multi-modal learning methods, existing methods still lack effective integration to fully explore synergies among spatio-temporal modalities effectively for gesture recognition. The problems are partially due to the fact that the existing manually designed network architectures have low efficiency in the joint learning of multi-modalities. In this paper, we propose the first neural architecture search (NAS)-based method for RGB-D gesture recognition. The proposed method includes two key components: 1) enhanced temporal representation via the proposed 3D Central Difference Convolution (3D-CDC) family, which is able to capture rich temporal context via aggregating temporal difference information; and 2) optimized backbones for multi-sampling-rate branches and lateral connections among varied modalities. The resultant multi-modal multi-rate network provides a new perspective to understand the relationship between RGB and depth modalities and their temporal dynamics. Comprehensive experiments are performed on three benchmark datasets (IsoGD, NvGesture, and EgoGesture), demonstrating the state-of-the-art performance in both single- and multi-modality settings. The code is available at  \href{https://github.com/ZitongYu/3DCDC-NAS}{https://github.com/ZitongYu/3DCDC-NAS}. 
\end{abstract}


\IEEEpeerreviewmaketitle

\section{Introduction}


\IEEEPARstart{A}{s} one of the multi-modal video understanding problems, RGB-D based video  gesture recognition~\cite{molchanov2016online,wan2016chalearn,ohn2014hand,zhang2018egogesture} has been applied to many real-world applications, such as virtual reality~\cite{weissmann1999gesture} and human-computer interaction~\cite{rautaray2015vision}. 3D convolutional neural network (3DCNN)~\cite{miao2017multimodal,li2016large,wang2017large,abavisani2019improving} and long short-term memory (LSTM)~\cite{zhang2018attention} have been adopted to learn the spatial-temporal features for gesture recognition. However, the learned spatio-temporal representation is still easily contaminated by irrelevant factors (e.g., illumination and background). A feasible solution is to add an extra enhanced temporal feature learning module~\cite{sun2018optical,lee2018motion,piergiovanni2019representation}, which is computational costly and tricky for the off-the-shelf 3DCNNs. \textbf{How to learn efficient spatio-temporal features in basic convolution operator for enriching temporal context is worth exploring for gesture recognition.}

As gestures have various temporal ranges, modeling such visual tempos would benefit for gesture recognition. Previous  methods~\cite{feichtenhofer2019slowfast,zhang2018dynamic,wang2017spatiotemporal} attempt to construct the frame pyramid for such purpose, with each branch of the frame pyramid sampling the input frames at a different rate. However, the architecture (i.e., network structure) of each branch and relations (i.e., lateral connections) among the multi-rate branches are usually shared and hand-designed, which is sub-optimal for message propagation. Hence, \textbf{how to discover better-suited architectures and lateral connections among multi-rate branches is crucial.}

For RGB-D based gesture recognition, complementary feature learning from different data modalities is beneficial. For example, the depth data is easy to distinguish foregrounds (i.e., face, hands, and arms) from backgrounds while RGB data provides detailed texture/color appearances. However, most existing  methods~\cite{miao2017multimodal,li2016large,neverova2015moddrop,pitsikalis2017multimodal,roitberg2019analysis} conduct the multi-modal fusion via coarse strategies (e.g, score fusion or last layer concatenation), which may leverage the multi-modal information insufficiently. Thus, \textbf{to design more reasonable multi-modal fusion strategy is not a trivial work.}

Motivated by the above observations, we propose a novel spatio-temporal convolution family called 3D Central Difference Convolution (3D-CDC), to exploit the rich local motion and enhance the spatio-temporal representation. Moreover, over the 3D-CDC-based enhanced search space, Neural Architecture Search (NAS) is adopted to automatically discover the optimized multi-rate and multi-modal networks for RGB-D gesture recognition. Our contributions include:

\begin{itemize}
    \item A novel spatio-temporal convolution family, 3D-CDC, is proposed, intending to capture rich temporal context via aggregating temporal difference information. Without introducing extra parameters, 3D-CDC can replace the vanilla 3D convolution, and plug and play in existing 3DCNNs for various modalities with enhanced temporal modeling capacity.  
    
    
    \item We propose a two-stage NAS method to automatically discover well-suited backbones and lateral connections for the multi-rate and multi-modal networks, which effectively explores RGB-D-temporal synergies and represents global dynamics.
    
    \item To our best knowledge, this is the first approach that searches multi-rate and multi-modal architectures for RGB-D gesture recognition. Our searched architecture provides a new perspective to understand the relationship among multi-rate branches as well as modalities.
    
    \item Our proposed method achieves state-of-the-art (SOTA) performance on three benchmark datasets on both single- and multi-modality testing.

\end{itemize}

In the rest of the paper, Sec.~\ref{sec:relatedwork} provides the related work. Sec.~\ref{sec:method} formulates the 3D-CDC family, and introduce the two-stage multi-rate and multi-modal NAS algorithm. Sec.~\ref{sec:experiment} provides rigorous ablation studies and evaluates the performance of the proposed models on three benchmark datasets. Sec.~\ref{sec:Visualization} shows the visualization results and discusses transferability to the action recognition task. Finally, a conclusion with future directions is given in Sec.~\ref{sec:conclusion}.

\section{Related Work}
\label{sec:relatedwork}

In this section, we first introduce some recent progress in multi-modal gesture recognition. Then, previous video-based NAS methods will be reviewed.

\noindent\textbf{Multi-Modal Gesture Recognition.}\quad  For video-based gesture recognition, it is challenging to track the motion of hands and arms owing to the large degree of freedom. Many hand-crafted feature based~\cite{wan20143d,wan2013one,malgireddy2012temporal,wan2015explore} and deep learning-based methods~\cite{ji2017spatial,liu2017continuous,zhang2018attention,simonyan2014two,wang2018cooperative} are proposed to tackle this issue. As for the learning-based gesture recognition, on one hand, 3DCNNs including C3D~\cite{tran2015learning}, Res3D~\cite{hara2018can}, I3D~\cite{carreira2017quo} and SlowFast~\cite{feichtenhofer2019slowfast} are utilized for gesture feature extractor. On the other hand, LSTM variants such as AttenConvLSTM~\cite{zhu2019redundancy,zhu2019redundancy} and PreRNN~\cite{yang2018making} are introduced for temporal memory propagation. Based on the CNNs, several extended modules~\cite{sun2018optical,lee2018motion,piergiovanni2019representation,diba2018spatio} and convolution operators~\cite{kumawat2019lp,yu2020searching} are developed for enhancing the spatio-temperal representation. However, most of them need extra structures and learnable parameters to modulate the original spatio-temporal features. In this paper, we propose 3D-CDC for modeling a rich temporal context, which is vital for describing fine-grained hands/arms motion. Among these methods, Lee et al.’s~\cite{lee2018motion} motion feature network (MFNet) and Yu et al.’s~\cite{yu2020searching,yu2020multi} central difference convolution (CDC) are the most similar to our work. Instead of the fixed motion filter used in MFNet and only the spatial context in CDC, our 3D-CDC learns the temporal gradient (motion) filters automatically in a unified 3D convolution operator.

Due to the development of the RGB-D cameras like Intel RealSense, RGB and depth modalities~\cite{li2016large,miao2017multimodal,narayana2018gesture} are favorite to complementarily fuse and improve the performance. Besides the RGB-D data, some other modalities such as optical flow~\cite{li2019large,miao2017multimodal}, depth flow~\cite{narayana2018gesture}, skeleton~\cite{liu20203d,liu20203ds} and saliency video~\cite{li2017large,duan2018unified,wang2017large,duan2018unified} are also employed for gesture recognition. However, the flow and saliency modalities need extra computational cost and might lose some vital information after pre-processing. 

In terms of the multi-modal fusion strategy, decision-level fusion~\cite{zhu2016large,wang2017large,zhang2017learning} and feature-level fusion~\cite{miao2017multimodal,li2016large,narayana2018gesture,roitberg2019analysis} methods have developed for integrating mutual knowledge from varied modalities. Despite achieving SOTA performance, the existing multi-modal fusion strategies for gesture recognition are designed manually and coarsely, which might be sub-optimal for message propagation between modalities. In this paper, we prefer to discover well-suitable multi-modal fusion strategies automatically via NAS.

\noindent\textbf{Video-based Neural Architecture Search.}\quad    
Our work is motivated by recent researches on NAS \cite{dong2019searching,liu2018darts,pham2018efficient,zoph2016neural,zoph2018learning,xu2019pc,yu2020auto}, while we focus on searching for multi-rate and multi-modal networks specially for RGB-D gesture recognition. The existing NAS methods could be summarized in three categories: 1) Reinforcement learning-based \cite{zoph2016neural,zoph2018learning}, 2) Evolution-based \cite{real2019regularized,real2017large}, and 3) Gradient-based \cite{liu2018darts,xu2019pc,cai2019proxylessnas}. From the perspective of NAS based video classification applications, single-modal based \cite{peng2019video,qiu2019scheduled,wang2020attentionnas} and multi-modal based \cite{perez2019mfas,ryoo2019assemblenet} methods have been developed for the action recognition task. Unlike AssembleNet ~\cite{ryoo2019assemblenet} which searches on RGB and optical flow modalities with single frame rate inputs, our work focuses on searching the synergies among multi-rate and multi-modal branches.

In terms of the search space design, cell-based NASNet \cite{zoph2018learning} space is favorite due to its flexibility and rich capacity. Besides, some novel operators, such as extended convolution \cite{yu2020searching,tan2019mixconv} and attention \cite{wang2020attentionnas,howard2019searching}, are introduced into search space, which is proved to be beneficial for searching more powerful architecture. However, there are still no operators specially designed for temporal enhancement.

To our best knowledge, no NAS based method has ever been proposed for RGB-D gesture recognition. To fill in the blank, we search multi-rate and multi-modal networks over the temporal enhanced search space for RGB-D gesture recognition.

\section{Methodology}
\label{sec:method}

In this section, we first introduce the 3D-CDC family in Sec.~\ref{sec:CDC}. Over the 3D-CDC based space, we then propose the two-stage multi-rate and multi-modal NAS in Sec.~\ref{sec:NAS}.

\subsection{ Temporal Enhancement via 3D-CDC}
\label{sec:CDC}



\begin{figure*}
\centering
\includegraphics[scale=0.20]{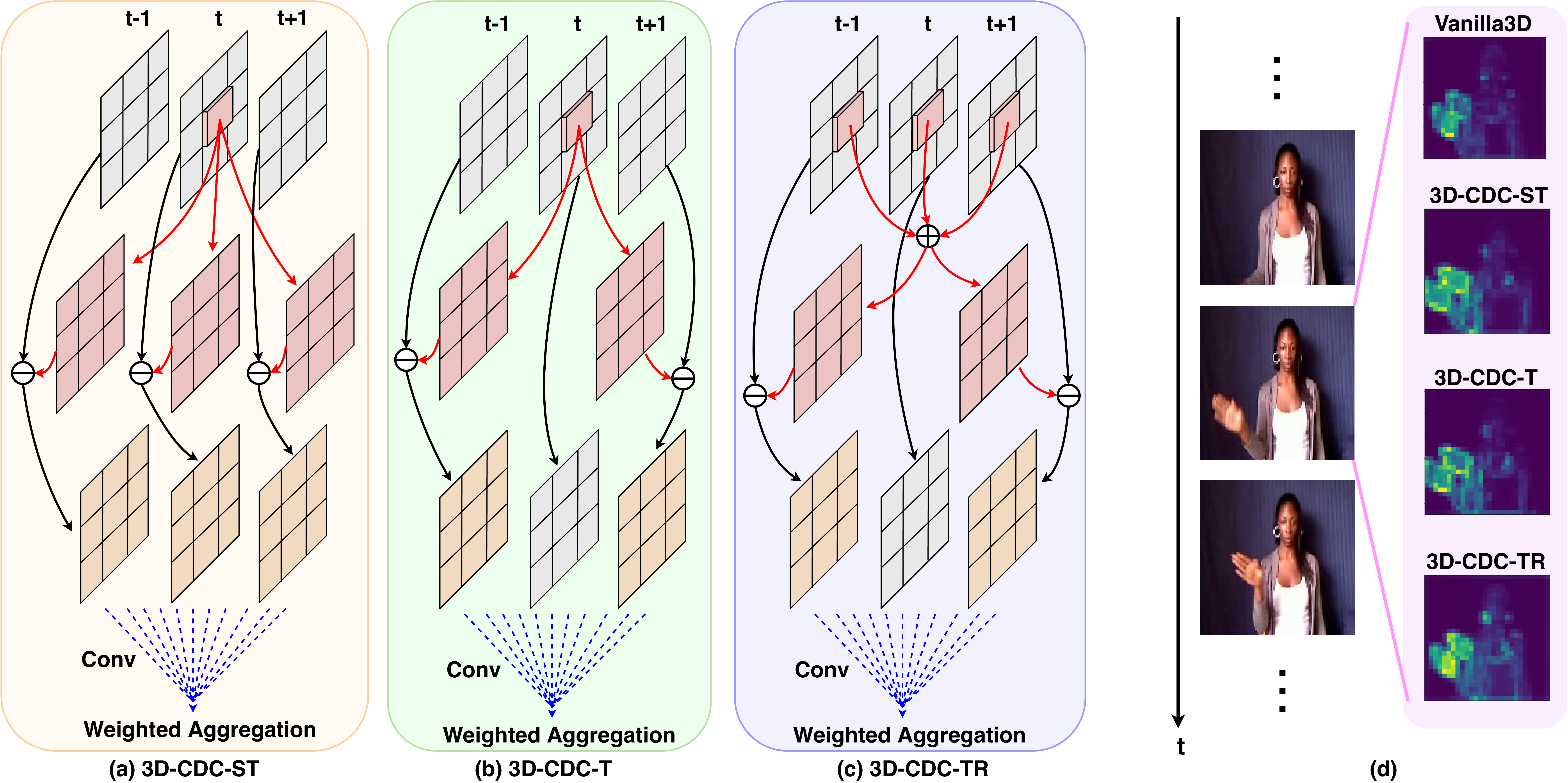}
  \caption{\small{
  Our proposed 3D-CDC family with kernel size 3, which could be used as novel operators for NAS. (a) 3D-CDC-ST considers the central difference information in the whole local spatio-temporal regions. (b) 3D-CDC-T only calculates the central difference clues from the local spatio-temporal regions of the adjacent frames. (c) 3D-CDC-TR is similar to 3D-CDC-T but adopts the temporal central mean pooling before calculating the central difference term, which is more robust to temporal noise. The symbols $\ominus$ and $\oplus$ denote element-wise subtraction and mean operations, respectively. (d) Feature response of various convolutions in RGB modality. Compared with vanilla 3D convolution, the 3D-CDC family enhances the temporal context obviously. 
  }
  }
\label{fig:CDC}
\end{figure*}

In classical 3DCNNs, 3D convolution is the most fundamental operator for spatio-temporal feature representation. In this subsection, for simplicity, the 3D convolutions are described in 3D (without channel) while an extension to 4D is straightforward. There are two main steps in the vanilla 3D convolution: 1) \textsl{sampling} local receptive field cube $\mathcal{C}$ over the input feature map $x$; 2) \textsl{aggregation} of sampled values via weighted summation with learnable weights $w$. Hence, the output feature map $Vanilla$ can be formulated as
\begin{equation} \small
Vanilla(p_0)=\sum_{p_n\in \mathcal{C}}w(p_n)\cdot x(p_0+p_n),
\label{eq:vanilla}
\end{equation}
where $p_0$ denotes current location on both input and output feature maps while $p_n$ enumerates the locations in $\mathcal{C}$. In this subsection, 3D convolution with kernel size 3$\times$3$\times$3 and dilation 1 is used for demonstration, and the other configurations are analogous. The local receptive field cube for the 3D convolution is $\mathcal{C}$=$\left \{  (-1,-1,-1),(-1,-1,0),\cdots,(0,1,1),(1,1,1) \right \}$.

\textbf{Spatio-Temporal Central Difference Convolution (3D-CDC-ST).}\quad Inspired by the CDC~\cite{yu2020searching} which introduces spatial gradient cues into representation learning, we integrate spatio-temporal gradient information into a unified 3D convolution operator. It is worth noting that such spatial gradient and temporal difference designs are widely used in the dense optical flow~\cite{brox2004high} calculation. Instead of the optical flow only performed in RGB sequence level, networks with stacked spatio-temporal CDC would be regularized to learn more local motion context in both RGB sequence and deep feature level, which is able to model fine-grained temporal dynamics for gesture recognition.

Similarly, spatio-temporal CDC also consists of two steps, i.e., \textsl{sampling} and \textsl{aggregation}. The sampling step is similar to vanilla 3D convolution but the aggregation step is different: as illustrated in Fig.~\ref{fig:CDC}(a), spatio-temporal CDC prefers to aggregate the center-oriented spatio-temporal gradient of sampled values. Eq.~(\ref{eq:vanilla}) becomes

\begin{equation} \small
\setlength{\belowdisplayskip}{-0.6em}
CDC(p_0)=\sum_{p_n\in \mathcal{C}}w(p_n)\cdot (x(p_0+p_n)-x(p_0)).
\label{eq:central}
\end{equation}
When $p_n$=(0,0,0), the gradient value always equals to zero with respect to the central location $p_0$ itself. For the gesture recognition task, both spatio-temporal intensity-level semantic information and gradient-level difference message are crucial and complementary. The former one is good at global modeling and robust to sensor-based noise while the latter one focuses more on local appearance and motion details and might be influenced by noise. As a result, combining vanilla 3D convolution with 3D-CDC might be a feasible manner to provide more robust and discriminative modeling capacity. Therefore we generalize spatio-temporal CDC as 

\begin{equation} \footnotesize
\setlength{\belowdisplayskip}{-0.1em}
\begin{split}
CDC_{ST}(p_0)
&=\theta \cdot CDC(p_0) + (1-\theta)\cdot Vanilla(p_0) \\
&=\underbrace{\sum_{p_n\in \mathcal{C}}w(p_n)\cdot x(p_0+p_n)}_{\text{vanilla 3D convolution}}+\theta\cdot (\underbrace{-x(p_0)\cdot\sum_{p_n\in \mathcal{C}}w(p_n))}_{\text{spatio-temporal CD term}}, \\
\label{eq:CDC-ST}
\end{split}
\end{equation} 
where hyperparameter $\theta \in [0,1]$ tradeoffs the contribution between intensity-level and gradient-level information. Please note that $w(p_n)$ is shared between vanilla 3D convolution and spatio-temporal central difference (CD) term, thus no extra parameters are added. 

\textbf{Temporal Central Difference Convolution (3D-CDC-T).}\quad Unlike the aforementioned spatio-temporal CDC considering both spatial and temporal gradient cues, we propose a version with only temporal central differences. As shown in Fig.~\ref{fig:CDC}(b), the sampled local receptive field cube $\mathcal{C}$ is separated into two kinds of regions: 1) the region in the current time step $\mathcal{R'}$, and 2) the regions in the adjacent time steps $\mathcal{R''}$. In the setting of a temporal CDC, the central difference term is only calculated from $\mathcal{R''}$. Thus the generalized temporal CDC can be formulated via modifying Eq.~(\ref{eq:CDC-ST}) as
\begin{equation} \footnotesize
\begin{split}
CDC_{T}(p_0)
&=\underbrace{\sum_{p_n\in \mathcal{C}}w(p_n)\cdot x(p_0+p_n)}_{\text{vanilla 3D convolution}}+\theta\cdot (\underbrace{-x(p_0)\cdot\sum_{p_n\in \mathcal{R''}}w(p_n))}_{\text{temporal CD term}}. \\
\label{eq:CDC-T}
\end{split}
\end{equation}

\begin{figure*}
\centering
\includegraphics[scale=0.36]{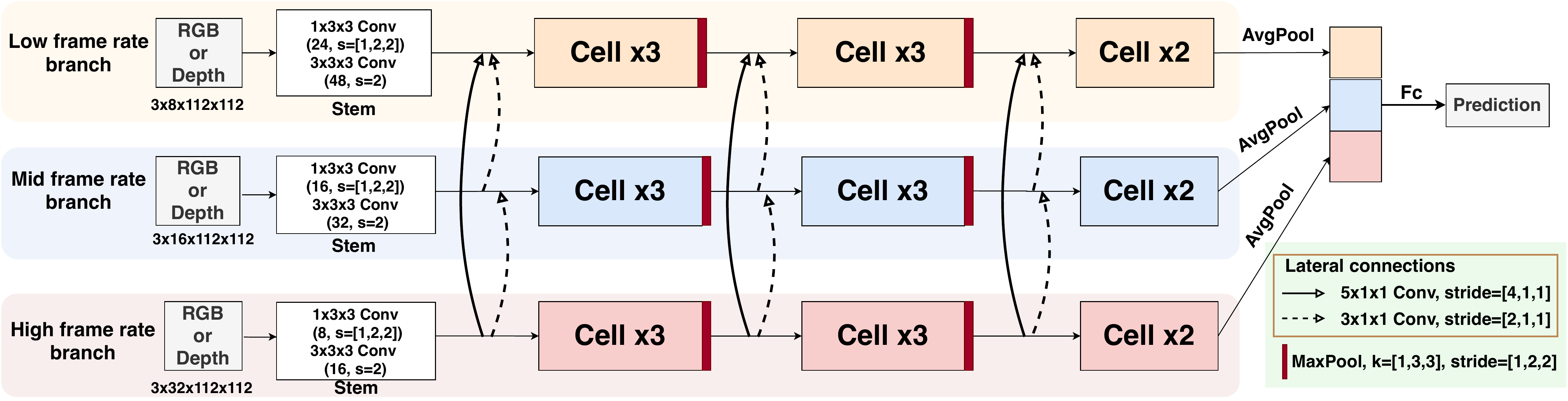}
  \caption{\small{
  Architecture search space in the first stage. Single-modal (RGB or depth) multi-rate frames are adopted as inputs. Here we utilize 3 branches with different frame rates (e.g., uniformly sampling the original videos into 8, 16, and 32 frames, respectively), and it also can be extended to more branches according to the actual situation. In this search stage, inspired by \cite{feichtenhofer2019slowfast}, the architectures of all lateral connections are fixed with temporal convolutions. And the outputs of the lateral connections are concatenated with the features from the target branch. The channel numbers are doubled after each MaxPool layer. The architecture of the cells from multi-rate branches to be searched can be shared or unshared (see Sec. \ref{sec:Ablation} for ablation study).
  }
  }

\label{fig:NAS1}
\end{figure*}

\textbf{Temporal Robust Central Difference Convolution (3D-CDC-TR).}\quad  In consideration of the sensor noise especially in the depth modality, we also propose a version with the temporal robust central difference. Similarly, the temporal robust CDC only calculates the difference term from the regions in the adjacent time steps $\mathcal{R''}$. As shown in Fig.~\ref{fig:CDC}(c), the robust temporal center is represented by averaging the spatial centers of all time steps (i.e., $p_0^{t-1}$, $p_0$ and $p_0^{t+1}$) within $\mathcal{C}$. The robust temporal center-oriented gradient might be less sensitive to the pixel jitters from the adjacent time steps. The generalized temporal robust CDC can also be formulated via modifying Eq.~(\ref{eq:CDC-ST}) as
\begin{equation} \footnotesize
\setlength{\belowdisplayskip}{-0.1em}
\begin{split}
CDC_{TR}(p_0)
&=\underbrace{\sum_{p_n\in \mathcal{C}}w(p_n)\cdot x(p_0+p_n)}_{\text{vanilla 3D convolution}} \\
&+\theta\cdot (\underbrace{-Avg[x(p_0^{t-1}),x(p_0),x(p_0^{t+1})]\cdot\sum_{p_n\in \mathcal{R''}}w(p_n))}_{\text{temporal robust CD term}}. \\
\label{eq:CDC-TR}
\end{split}
\end{equation}

We will henceforth refer to these three generalized versions (i.e., Eq.~(\ref{eq:CDC-ST}), (\ref{eq:CDC-T}) and (\ref{eq:CDC-TR})) as 3D-CDC-ST, 3D-CDC-T and 3D-CDC-TR, respectively. The ablation studies about the 3D-CDC family and hyperparameter $\theta$ are conducted in Sec.~\ref{sec:Ablation}.

\textbf{Relation between 3D-CDC and 3D Vanilla Convolution.}\quad  Compared to 3D vanilla convolution, 1) 3D-CDC-ST regularizes the spatio-temporal representation with more detailed spatial cues and temporal dynamics, which might be suitable for scene-aware video understanding tasks; 2) 3D-CDC-T enhances the spatio-temporal representation with only rich temporal context, which might be assembled for video temporal reasoning tasks; and 3) 3D-CDC-TR introduces robust but slighter temporal evolution cues for spatio-temporal representation, which might perform robustly even in noisy scenarios. In particular, 3D-CDC-ST, 3D-CDC-T, and 3D-CDC-TR degrade to vanilla 3D convolution when $\theta$=0. As illustrated in Fig.~\ref{fig:CDC}(d), the 3D-CDC family provides more details about the trajectory of the left arm, and such a local motion context is crucial to gesture recognition. More visualizations are shown in Sec.~\ref{sec:Visualization2}.

\subsection{Multi-Rate and Multi-Modal NAS}
\label{sec:NAS}

In order to seek the best-suited backbones and lateral connections for multi-rate and multi-modal networks, we propose a two-stage NAS method to 1) search backbones for multi-rate single-modal networks first, and then 2) search lateral connections for multi-modal networks based on the searched backbones. The iterative procedure is outlined in Algorithm \ref{algorithm:Auto-Label}. More technical details can be referred to two gradient-based NAS methods \cite{liu2018darts,xu2019pc}.

\begin{algorithm}[t]\small
		\caption{Two-Stage Multi-Rate and Multi-Modal NAS}
		\textit{{\bfseries Stage1:} Fix lateral connections, and search backbones} 

		For multi-rate backbones, create a mixed operation $\tilde{o}_{b}^{(i,j)}$ parametrized by $\alpha_{b}^{(i,j)}$ for each edge $(i,j)$ \\
		{ \small{1}\,\,\,:} {\bfseries for } each of modalities $\mathcal{M}$ {\bfseries do}  \\
		{ \small{2}\,\,\,:} \ \ \ \,Fix the lateral connections among multi-rate branches \\
		{ \small{3}\,\,\,:} \ \ \ \,{\bfseries while } not converged {\bfseries do} \\
		{ \small{4}\,\,\,:} \ \ \ \, \ \ \ \,Update architecture $\alpha_{b}$ by descending $\nabla_{\alpha_{b}}\mathcal{L}_{val}(\Phi_{b},\alpha_{b})$  \\
		{ \small{5}\,\,\,:} \ \ \ \, \ \ \ \,Update weights $\Phi_{b}$ by descending $\nabla_{\Phi_{b}}\mathcal{L}_{train}(\Phi_{b},\alpha_{b})$ \\
		{ \small{6}\,\,\,:} \ \ \ \,	{\bfseries end}\\
		{ \small{7}\,\,\,:} \ \ \ \,Derive the final backbone of the current modality based on the learned $\alpha_{b}$ \\
		{ \small{8}\,\,\,:} {\bfseries end}
		
		\textit{{\bfseries Stage2:} Fix backbones, and search lateral connections}
		
		For lateral connections, create a mixed operation $\tilde{o}_{c}^{(i,j)}$ parametrized by $\alpha_{c}^{(i,j)}$ for each edge $(i,j)$
	    
	    { \small{9}\,\,\,:} Initialize and fix the multi-rate and multi-modal backbones searched in \textit{\bfseries Stage 1} \\
		{ \small{10}\,\,\,:} {\bfseries while } not converged {\bfseries do} \\
		{ \small{11}\,\,\,:} \ \ \ \,Update architecture $\alpha_{c}$ by descending $\nabla_{\alpha_{c}}\mathcal{L}_{val}(\Phi_{c},\alpha_{c})$  \\
		{ \small{12}\,\,\,:} \ \ \ \,Update weights $\Phi_{c}$ by descending $\nabla_{\Phi_{c}}\mathcal{L}_{train}(\Phi_{c},\alpha_{c})$ \\
		{ \small{13}\,\,\,:} {\bfseries end}\\
		{ \small{14}\,\,\,:} Derive the final lateral connections based on the learned $\alpha_{c}$ 
		\label{algorithm:Auto-Label}
	
	\end{algorithm}

\begin{figure*}[t]
\centering

\includegraphics[scale=0.32]{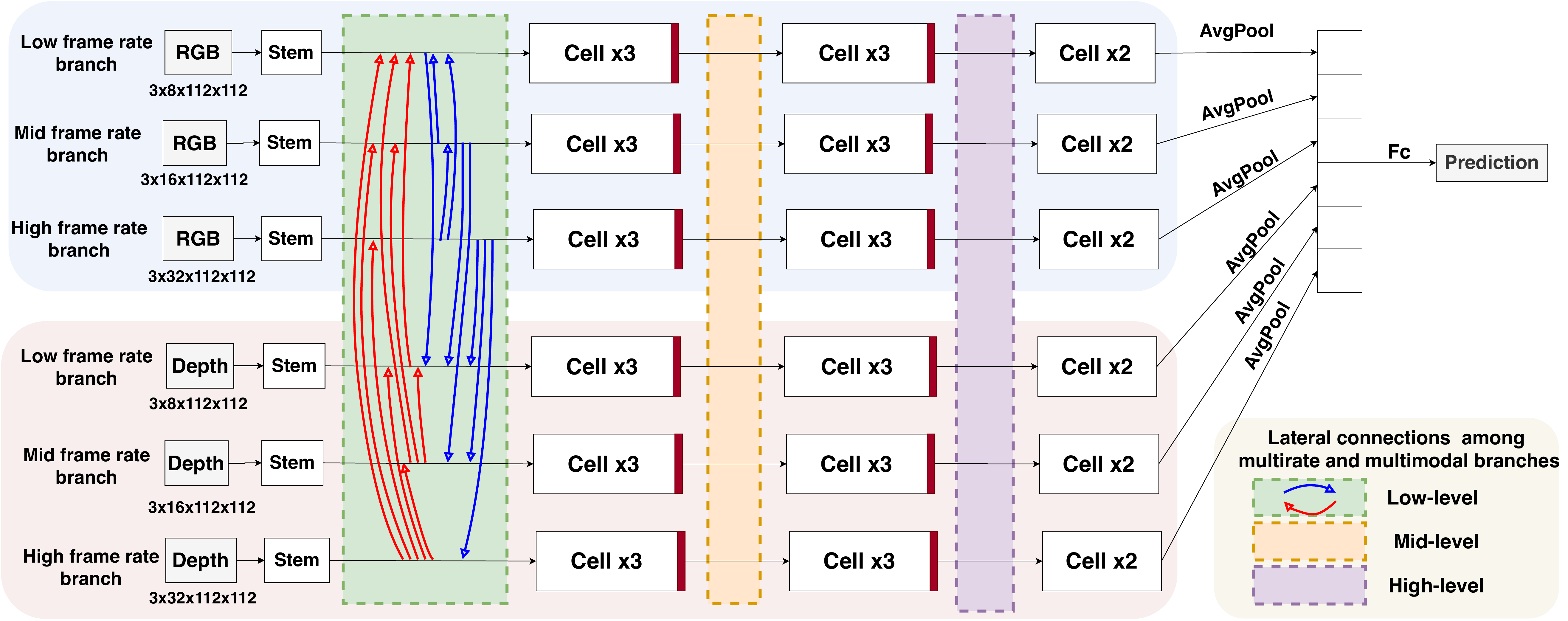}
  \caption{\small{
  Architecture search space in the second stage. Multi-modal and multi-rate frames are adopted as inputs. Here we utilize 3 branches with different frame rates for two modalities (RGB and depth), respectively. In this search stage, all cells are initialized with the searched structures in the first stage and then fixed. The architecture of the low-, mid- and high-level lateral connections to be searched can be shared or unshared.}
  }
 
\label{fig:NAS2}
\end{figure*}

\subsubsection{Stage 1: Searching Backbones for Multi-Rate Single-Modal Networks}

In SlowFast Networks \cite{feichtenhofer2019slowfast}, low- and high-rate branches are utilized for complementarily capturing dynamic visual tempos. However, the coarse hand-defined architecture with vanilla convolutions limits its representation capacity. Here we search optimal rate-aware backbones over the temporal enhanced search space. 

As illustrated in Fig.~\ref{fig:NAS1}, in the first stage, our goal is to search for cells to form multi-rate backbones for single-modal gesture recognition. As for the cell-level structure, similar to \cite{liu2018darts}, each cell is represented as a directed acyclic graph (DAG) of $K$ nodes $\left \{ n \right \}^{K-1}_{i=0}$, where each node represents a network layer. We denote the operation space as $\mathcal{O}_{b}$, which consists of seven designed candidate operations: `\textit{Zero}', `\textit{Identity}', `\textit{Conv\_1x3x3}',  `\textit{CDC-T-06\_3x1x1}', `\textit{CDC-T-06\_3x3x3}', `\textit{CDC-TR-03\_3x1x1}' and `\textit{CDC-TR-03\_3x3x3}'. To be specific, `\textit{CDC-T-06}' and `\textit{CDC-TR-03}' denote the 3D-CDC-T with $\theta$=0.6 and 3D-CDC-TR with $\theta$=0.3, respectively. These settings of $\theta$ are based on the ablation study results in Section~\ref{sec:Ablation}. We also consider a vanilla operation space with vanilla 3D convolutions instead of 3D-CDC for comparison.


The multi-rate backbones for each of modalities $\mathcal{M}$ to be searched have the architecture parameters $\alpha^{(i,j)}_{b}$. Each edge $(i,j)$ of DAG represents the information flow from node $n_{i}$ to node $n_{j}$, which consists of the candidate operations weighted by the architecture parameter $\alpha^{(i,j)}_{b}$. Specially, each edge $(i,j)$ can be formulated by a function $\tilde{o_{b}}^{(i,j)}$ where $\tilde{o_{b}}^{(i,j)}(n_i)=\sum_{o_{b}\in \mathcal{O}_{b}}\eta_{o_{b}}^{(i,j)}\cdot o_{b}(n_{i})$. Softmax $\eta_{o_{b}}^{(i,j)}=\frac{exp(\alpha_{o_{b}}^{(i,j)})}{\sum_{{o}'_{b}\in \mathcal{O_{b}}}exp(\alpha_{{o}'_{b}}^{(i,j)})}$ is utilized to relax architecture parameter $\alpha^{(i,j)}_{b}$ into operation weight $o_{b}\in \mathcal{O}_{b}$. The intermediate node can be denoted as $n_{j}=\sum_{i<j}{\tilde{o_{b}}}^{(i,j)}(n_{i})$. The output node $n_{K-1}$ is depth-wise concatenation of all the intermediate nodes excluding the input nodes.


In the searching stage, cross-entropy loss is utilized for the training loss $\mathcal{L}_{train}$ and validation loss $\mathcal{L}_{val}$. Network parameters $\Phi_{b}$ and architecture parameters $\alpha_{b}$ are learned via solving the bi-level optimization problem:
\begin{equation} 
\begin{split}
&\underset{\alpha_{b}}{min} \quad \mathcal{L}_{val}(\Phi_{b} ^{*}(\alpha_{b}),\alpha_{b} ), \\
&s.t. \quad \Phi_{b} ^{*}(\alpha)=arg  \,
\underset{\Phi_{b} }{min}   \;
\mathcal{L}_{train}(\Phi_{b} ,\alpha_{b}).
\end{split}
\label{eq:optimize}
\end{equation}  
When the searching is converged, we derive the final architectures via: 1) setting $o_{b}^{(i,j)}=arg\,max_{o_{b}\in \mathcal{O}_{b},o_{b}\neq zero}\,\eta_{o_{b}}^{(i,j)}$, and 2) for each intermediate node, choosing two incoming edges with the two largest values of $max_{o_{b}\in \mathcal{O}_{b},o\neq zero}\,\eta_{o_{b}}^{(i,j)}$.

\subsubsection{Stage 2: Searching Lateral Connections for Multi-Rate Multi-Modal Networks}

The lateral connections from most existing multi-rate \cite{feichtenhofer2019slowfast} or multi-modal \cite{wang2017large,miao2017multimodal,roitberg2019analysis} spatio-temporal networks are designed manually, which might be sub-optimal for information exchange. Here we propose to search best-suited lateral connections among rate-aware and modality-aware branches, intending to effectively explore RGB-D-temporal synergies. In the second stage, our goal is to search for lateral connections among the multi-rate and multi-modal branches. As most of the definitions and the search procedure are similar to those in the first stage, here we only show the two main differences from the first stage.

On one hand, the composition of architecture search space is different. As shown in Fig.~\ref{fig:NAS2} (see low-level connections for example), the lateral connections search space can be represented as a bidirectional graph of $K'$=6 nodes (branches) within the modalities $\mathcal{M}$. Specially, the lateral connections from the lower frame rate branches to the higher ones are not considered because we assume that the lower frame rate branches always have less information than the higher ones. Thus, there are total 18 edges (lateral connections) inside the bidirectional graph and each edge consists of the candidate operations weighted by its corresponding architecture parameter $\alpha_{c}$. The final output of each node is the depth-wise concatenation of all outputs of the incoming edges.

On the other hand, the design of the operation space is different. The operation space for lateral connections is denoted as $\mathcal{O}_{c}$, which consists of two parts: 1) `\textit{Zero}', `\textit{Conv\_5x1x1}',  `\textit{CDC-T-06\_5x1x1}', `\textit{CDC-TR-03\_5x1x1}' with $stride$=(4,1,1) for the edges from the high frame rate branches to the low frame rate branches; and 2) `\textit{Zero}', `\textit{Conv\_3x1x1}',  `\textit{CDC-T-06\_3x1x1}', `\textit{CDC-TR-03\_3x1x1}' with $stride$=(2,1,1) for the others. Specially, $stride$=1 is utilized for edges between the branches of different modalities with same frame rate. When the search is converged, for each edge $(i,j)$, only the operation in $\mathcal{O}_{c}$ with the largest $\alpha^{(i,j)}_{c}$ is adopted. With the two-stage multi-rate and multi-modal NAS in Algorithm \ref{algorithm:Auto-Label}, both the final backbones and lateral connections are derived.

\section{Experiments}
\label{sec:experiment}

In this part, we first give details for benchmark datasets and experimental setup. Then, we thoroughly evaluate the impacts of 3D-CDC family, multi-rate configuration, and two-stage NAS. Finally, we evaluate and compare our results with state-of-the-art methods on three benchmark datasets.

\vspace{-0.8em}
\subsection{Datasets}
\label{sec:Datasets}
We evaluate our method on three widely used RGB-D gesture datasets: Chalearn IsoGD \cite{wan2016chalearn,wan2019chalearn}, NVGesture \cite{molchanov2016online} and EgoGesture \cite{zhang2018egogesture} datasets. The \textbf{Chalearn IsoGD dataset} \cite{wan2016chalearn} contains 47,933 RGB-D gesture videos divided into 249 kinds of gestures and is performed by 21 individuals. The dataset contains 35878 training, 5784 validation, and 6271 testing samples. We follow the SOTA methods \cite{zhang2017learning,zhu2019redundancy,narayana2018gesture} to evaluate performance on the validation set. The \textbf{NVGesture dataset} \cite{molchanov2016online} focuses on touchless driver controlling. It contains 1532 dynamic gestures fallen into 25 classes. It includes 1050 samples for training and 482 for testing. The videos are recorded with three modalities (RGB, depth, and infrared). For fair evaluations with SOTA methods, infrared modality is not used in our experiments. The \textbf{EgoGesture dataset} \cite{zhang2018egogesture} is a large multi-modal egocentric hand gesture dataset. It contains 24,161 hand gesture clips of 83 classes of gestures, performed by 50 subjects. Videos in this dataset are captured with an Intel RealSense SR300 device in RGB-D modalities across multiple indoor and outdoor scenes.

\vspace{-0.8em}
\subsection{Implementation Details}
\label{sec:Details}
Our proposed method is implemented with Pytorch. Cell nodes $K$=4 and $K'$=6 are used as the default setting. The optical flow is extracted by pyflow \cite{pathak2017learning} - a python wrapper for dense optical flow \cite{brox2004high}. \textbf{In the search phase}, partial channel connection and edge normalization \cite{xu2019pc} are adopted. The initial channel numbers for low, mid, and high frame rate branches are 24, 16, and 8, respectively, which double after searching. There are 8 cells for each branch in the search stage, which increases to 12 cells after searching. SGD optimizer with learning rate lr=1e-2 and weight decay wd=5e-5 is utilized when training the network weights. The architecture parameters are trained with Adam optimizer with lr=6e-4 and wd=1e-3. The lr decays with factor 0.5 in the 20$^{th}$ epoch. We search 30 epochs on the training set of IsoGD dataset \cite{wan2016chalearn} with batch size 20 while architecture parameters are not updated in the first 10 epochs. Especially, $\mathcal{L}_{train}$ is calculated on the first half of the training set while $\mathcal{L}_{val}$ on the latter part. The whole two-stage NAS costs 12 days on four P100s. \textbf{In the training phase}, models are trained with SGD optimizer with initial lr=1e-2 and wd=5e-5. The lr decays with factor 0.1 when the validation accuracy has not improved within 3 epochs. Random horizontal flip and spatial crops are utilized for data augmentation. We train models with batch size 12 for maximum 80 epochs on four RTX-2080Ti GPUs. 


\vspace{-0.8em}
\subsection{Ablation Study}
\label{sec:Ablation}
All ablation studies are trained from scratch and evaluated on the validation set of the IsoGD dataset.

\begin{figure}[t]
\includegraphics[scale=0.22]{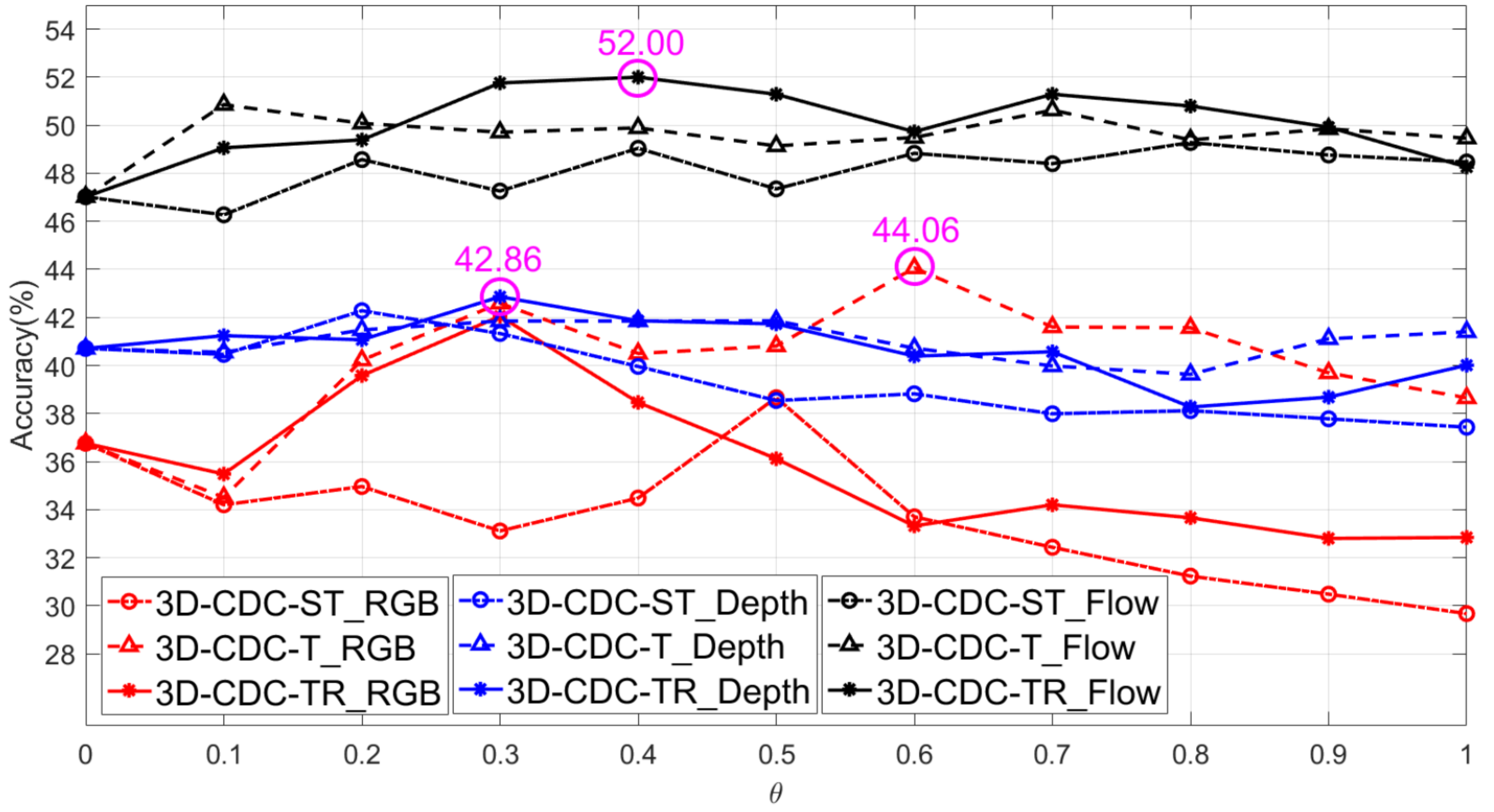}
 \caption{\small{
  Impact of 3D-CDC for RGB, depth and optical flow.}
  }
 
\label{fig:Ablation1}
\end{figure}

\textbf{Impact of 3D-CDC for Modalities.}\quad 
In these experiments, we use C3D \cite{tran2015learning} as the backbone and sequence size 3$\times$16$\times$112$\times$112 as the inputs. According to Eq.~(\ref{eq:CDC-ST}), (\ref{eq:CDC-T}) and (\ref{eq:CDC-TR}), $\theta$ controls the contribution of the temporal difference cues. As illustrated in Fig.~\ref{fig:Ablation1}, 3D-CDC-T improves the accuracy of RGB modality dramatically. Compared with vanilla 3D convolution (i.e., $\theta$=0), 3D-CDC-T gains 7.3\% when $\theta$=0.6, which indicates the advantages of temporal difference context. One highlight is that, assembled with 3D-CDC-T, the RGB modality is able to obtain comparable accuracy (44.06\% vs. 47.02\%) with optical flow modality, indicating an excellent dynamic modeling capacity of 3D-CDC-T. In terms of the depth and optical flow modalities, the best performance (42.86\% and 52\%) could be achieved when using 3D-CDC-TR with $\theta$=0.3 and $\theta$=0.4, respectively. Compared with 3D-CDC-T, 3D-CDC-TR is more robust for depth and optical flow modalities because it alleviates sensor noises and pre-processing artifacts between frames in these two modalities. By the observation, the 3D-CDC-ST performs relatively poorly. The reason might be that the gesture recognition task prefers more temporal reasoning context than spatial gradient cues and appearance details. According to their enhanced temporal representation abilities, 3D-CDC-T with $\theta$=0.6 and 3D-CDC-TR with $\theta$=0.3 are considered in our NAS operation space.

\begin{figure}[t]
\centering
\vspace{-0.4em}
\includegraphics[scale=0.57]{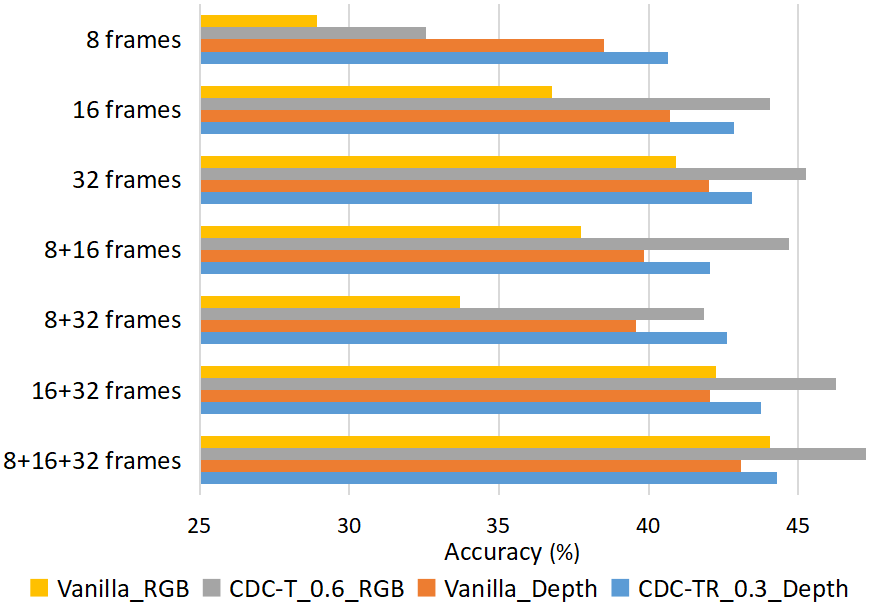}
\vspace{-1.4em}
  \caption{\small{
  The ablation study of the multi-rate networks. `8+16+32 frames' means that there are three branches with different frame rate (i.e., temporally downsampling to 8, 16 and 32 frames, respectively) as inputs. `CDC-T\_0.6\_RGB' and `CDC-TR\_0.3\_Depth' denotes using 3D-CDC-T with $\theta$=0.6 for RGB inputs and 3D-CDC-TR with $\theta$=0.3 for depth inputs, respectively.}
  }
 
\label{fig:multirate}
\vspace{-0.3em}
\end{figure}

\textbf{Impact of Multi-Rate Branches.}\quad  As gestures have various temporal scales, modeling such visual tempos of different gestures facilitates their recognition. Here we conduct the ablation study with C3D \cite{tran2015learning} network to explore how the branches with different frame rates influence the gesture recognition task. As illustrated in Fig.~\ref{fig:multirate}, for the single rate network, the higher the frame rate it has, the better performance it will be. This is because a higher frame rate usually has less sampling temporal information loss, which has richer fine-grained temporal cues for gesture recognition. Furthermore, we could find that the performance could be further improved when cooperated with the multi-rate branches (e.g., `16 + 32 frames' and `8 + 16 + 32 frames'). In terms of the impact of multi-rate branches for different modalities, it is obvious that multi-rate branches contribute more to RGB than depth modality. When assembling with 3D-CDC-T for RGB or 3D-CDC-TR for depth, the trends of multi-rate networks are analogous as the vanilla cases but achieving holistic performance gains (due to the excellent representation capacity of 3D-CDC). 

We also reimplement SlowFast \cite{feichtenhofer2019slowfast} Networks (trained from scratch) with `8+32 frames' multi-rate setting on the IsoGD dataset. However, it only achieves respective 22.28\% and 40.69\% accuracy on RGB and depth inputs, which indicates the importance of suitable architecture design for multi-rate networks in the gesture recognition task.

\begin{table}[t]
\centering
\caption{Comparison among various configurations of the two-stage NAS for varied modalities. The upper part is about the first stage NAS1 while bottom part is about the second stage NAS2. The evaluation metric is accuracy (\%).}
\resizebox{0.46\textwidth}{!}{
\begin{tabular}{c c c c}

 \toprule
 Configuration & RGB & Depth & RGB-D \\
 \midrule

w/o CDC, w/o NAS & 44.07 & 43.10 & -\\
w/o CDC, w/ NAS1 & 44.86 & 43.57 & -\\
w/ CDC, w/o NAS & 47.26 & 44.28 & -\\
w/ CDC, w/ NAS1 & 48.38 & 45.08 & -\\

\midrule
w/ CDC, w/ NAS2\_{Fixed} & - & - & 42.62 \\
w/ CDC, w/ NAS2\_{Shared} & - & - & 46.43 \\
w/ CDC, w/ NAS2\_{Unshared} & - & - & 50.17 \\

 \bottomrule
\end{tabular}
}
\label{tab:NAS}
\end{table}

\textbf{Effectiveness of the Two-stage NAS.}\quad 
Based on the best multi-rate setting (`8+16+32 frames'), we study the two-stage NAS for both single and multiple modalities. The first stage NAS (\textbf{NAS1}) intends to find well-suited multi-rate single-modal networks. As illustrated in Tab.~\ref{tab:NAS}, when searching over the vanilla search space w/o CDC, the architectures found by NAS1 improve 0.79\% and 0.47\% accuracy (compared with multi-rate C3D) for RGB and depth inputs, respectively. Moreover, the gains consistently occur when searching over 3D-CDC based search space for both RGB (+1.12\%) and depth (+0.8\%) modalities.


Based on the searched multi-rate networks for RGB and depth modalities, `NAS2\_Fixed' utilizes late fusion directly without searching the lateral connections between two modalities. Out of expectation, it performs even worse than the single-modal NAS1 searched models. It means that simply late fusion will encounter the problem of insufficient information exchange in feature levels. With the second stage NAS (\textbf{NAS2}), `NAS2\_Unshared' achieves more than 50\% accuracy, which indicates the advantages of NAS that mines 
the efficient integration of multi-rate and multi-modal branches. Furthermore, compared with `NAS2\_Shared' searching the shared lateral connections for low-mid-high levels, `NAS2\_Unshared' performs better (+3.74\%), which implies the importance of the specific design for interactions of each level.   


\begin{table}[t]
\centering
\caption{Results on the validation set of IsoGD \cite{wan2016chalearn}. }
\resizebox{0.47\textwidth}{!}{
\begin{tabular}{c c c}

 \toprule
 Method & Modality & Accuracy (\%) \\
 \midrule
Hu \etal \cite{Hu2018Learning}  &RGB & 44.88 \\
Res3D  \cite{li2016large} &RGB &45.07  \\
Zhang  \etal \cite{zhang2017learning}  & RGB &51.31 \\
Zhang  \etal \cite{zhang2018attention}  & RGB & 55.98 \\
Zhu  \etal \cite{zhu2019redundancy}  & RGB & 57.42 \\
 \textbf{NAS1 (Ours)} &RGB &\textbf{58.88} \\ 
 \midrule
Narayana   \etal \cite{narayana2018gesture} &Depth &27.98 \\
Res3D \cite{li2016large}  &Depth &48.44 \\
Zhang  \etal \cite{zhang2017learning} & Depth &49.81 \\
Zhu  \etal \cite{zhu2019redundancy}  & Depth & 54.18 \\
 \textbf{NAS1 (Ours)} &Depth &\textbf{55.68} \\
 \midrule
 
Zhu \etal \cite{zhu2017multimodal}  & RGB-D & 51.02  \\
Hu  \etal \cite{Hu2018Learning}   &RGB-D & 54.14\\
Zhang \etal \cite{zhang2017learning} &RGB-D & 55.29 \\
Zhu  \etal \cite{zhu2019redundancy}  & RGB-D & 61.05 \\
\textbf{NAS2 (Ours)} & RGB-D &60.04  \\
\textbf{NAS1+NAS2 (Ours)} & RGB-D & \textbf{65.54} \\
 \midrule
Li  \etal \cite{li2019large}   &  RGB-D-Flow &54.50 \\
Zhang  \etal \cite{zhang2017learning} & RGB-D-Flow &58.65  \\

Miao  \etal \cite{miao2017multimodal}  &RGB-D-Flow &64.40 \\


\textbf{NAS2 (Ours)} & RGB-Flow &61.22  \\
\textbf{NAS2 (Ours)} & Flow-Depth &62.47  \\
\textbf{NAS2\_all (Ours)} & RGB-D-Flow & \textbf{66.23}  \\

 \midrule

FOANet w/o hand\cite{narayana2018gesture}  &RGB-D-Flow-DFlow &61.40  \\
FOANet w/ hand \cite{narayana2018gesture}  &RGB-D-Flow-DFlow &\textbf{80.96}  \\

 \bottomrule
\end{tabular}
}
\label{tab:IsoGD}
\end{table}

\vspace{-0.8em}
\subsection{Comparison with State-of-the-art Methods}
\label{sec:SOTA}

After studying the components in Sec.~\ref{sec:Ablation}, we evaluate our models on three benchmark datasets. Note that in this subsection, our models are firstly pre-trained on the Jester \cite{materzynska2019jester} gesture dataset, which is similar to \cite{zhang2017learning,zhu2019redundancy,kopuklu2019real}. 

\textbf{Results on IsoGD.} \quad  
As shown in Table~\ref{tab:IsoGD}, although the existing methods \cite{li2016large,zhang2017learning,zhang2018attention} adopt 3DCNNs to learn from single RGB or depth modality, it is still challenging to represent the discriminative and robust spatio-temporal features with vanilla 3D convolutions and coarsely designed architecture. With the enhanced temporal representation capacity via 3D-CDC and multi-rate collaboration, our proposed multi-rate single-modal NAS method `NAS1' obtains the best accuracy on every single modality. This exactly demonstrates the superiority of the searched architecture. In terms of the RGB-D gesture recognition, our searched architecture with two-stage NAS (NAS2) obtains more than 1\% and 4\% accuracy gains compared with the `NAS1' with RGB and Depth modality, respectively. It demonstrates the effectiveness of RGB-D-temporal synergies at earlier stages. Similar to \cite{zhu2019redundancy} ensembling the results from varied modalities, our `NAS1+NAS2' boosts the accuracy to 65.54\%. 

To evaluate the modality generalization of the architecture searched from RGB-D, we retrain the same model `NAS2' with RGB-Flow and Flow-Depth modalities separately and also achieve comparable performance (61.22\% and 62.47\%, respectively). To our best knowledge, it is the first to explore the modality generalization issues for multi-modal NAS. Finally, the best accuracy (66.23\%) could be achieved when ensembling the scores from all three `NAS2' models among RGB-D-Flow modalities. Although the FOANet \cite{narayana2018gesture}  reports the best performance (80.96\%) on IsoGD, the high accuracy is achieved by fusing 12 channels (i.e., global/left/right channels for four modalities) with manual hand detection. Note that without hand detection preprocessing, our `NAS2\_all' outperforms FOANet (66.23\% vs. 61.4\%) by a large margin using only RGB-D-Flow modalities. This exactly demonstrates the superiority of our searched architectures.

\begin{table}[t]
\centering
\caption{Results on the NVGesture \cite{molchanov2016online} dataset.}
\resizebox{0.39\textwidth}{!}{
\begin{tabular}{c c c}

 \toprule
 Method & Modality & Accuracy (\%) \\
 \midrule
HOG+HOG$^2$ \cite{ohn2014hand}  &RGB & 24.50 \\
Simonyan \etal \cite{simonyan2014two} &RGB & 54.60 \\
Wang \etal \cite{wang2016robust}  &RGB & 59.10 \\
C3D \cite{tran2015learning} &RGB & 69.30 \\
R3DCNN \cite{molchanov2016online}  &RGB & 74.10 \\
GPM \cite{gupta2019progression}  &RGB & 75.90 
\\
PreRNN \cite{yang2018making}  &RGB & 76.50 \\
ResNeXt-101 \cite{kopuklu2019real} &RGB & 78.63 \\
MTUT \cite{abavisani2019improving}  &RGB & 81.33 \\

\textbf{NAS1 (Ours)} &RGB &\textbf{83.61}\\
\midrule
HOG+HOG$^2$ \cite{ohn2014hand}  &Depth & 36.30 \\
SNV \cite{yang2014super}  &Depth &70.70 \\
C3D \cite{tran2015learning}  &Depth & 78.80 \\
R3DCNN \cite{molchanov2016online} &Depth & 80.30 \\
ResNeXt-101 \cite{kopuklu2019real}  &Depth & 83.82 \\
PreRNN \cite{yang2018making} &Depth & 84.40 \\
MTUT \cite{abavisani2019improving}  &Depth & 84.85 \\
GPM \cite{gupta2019progression}  &Depth & 85.50 
\\

\textbf{NAS1 (Ours)}  &Depth &\textbf{86.10}\\
\midrule
HOG+HOG$^2$ \cite{ohn2014hand} &RGB-D & 36.90 \\
I3D \cite{carreira2017quo} &RGB-D & 83.82 \\
PreRNN \cite{yang2018making} &RGB-D & 85.00 \\
MTUT \cite{abavisani2019improving}  &RGB-D & 85.48 \\
GPM \cite{gupta2019progression}  &RGB-D & 86.10 
\\
\textbf{NAS2 (Ours)}  &RGB-D &86.93 \\
\textbf{NAS1+NAS2 (Ours)}  &RGB-D &\textbf{88.38} \\
 \bottomrule
\end{tabular}
}
\label{tab:NVGesture}
\end{table}

\textbf{Results on NVGesture.} \quad   Table~\ref{tab:NVGesture} compares the performance of our method with SOTA methods on the NVGesture dataset. It can be seen that our approach performs the best for both single-modal and multi-modal testing, which indicates 1) our searched architecture is able to represent discriminative spatio-temporal features for single/multi-modal gesture recognition; and 2) the architecture searched on the source dataset (IsoGD) via the proposed two-stage NAS transfers well on the target dataset (NVGesture), demonstrating the excellent generalization ability of the proposed NAS method. 

\begin{figure}
\centering

\includegraphics[scale=0.38]{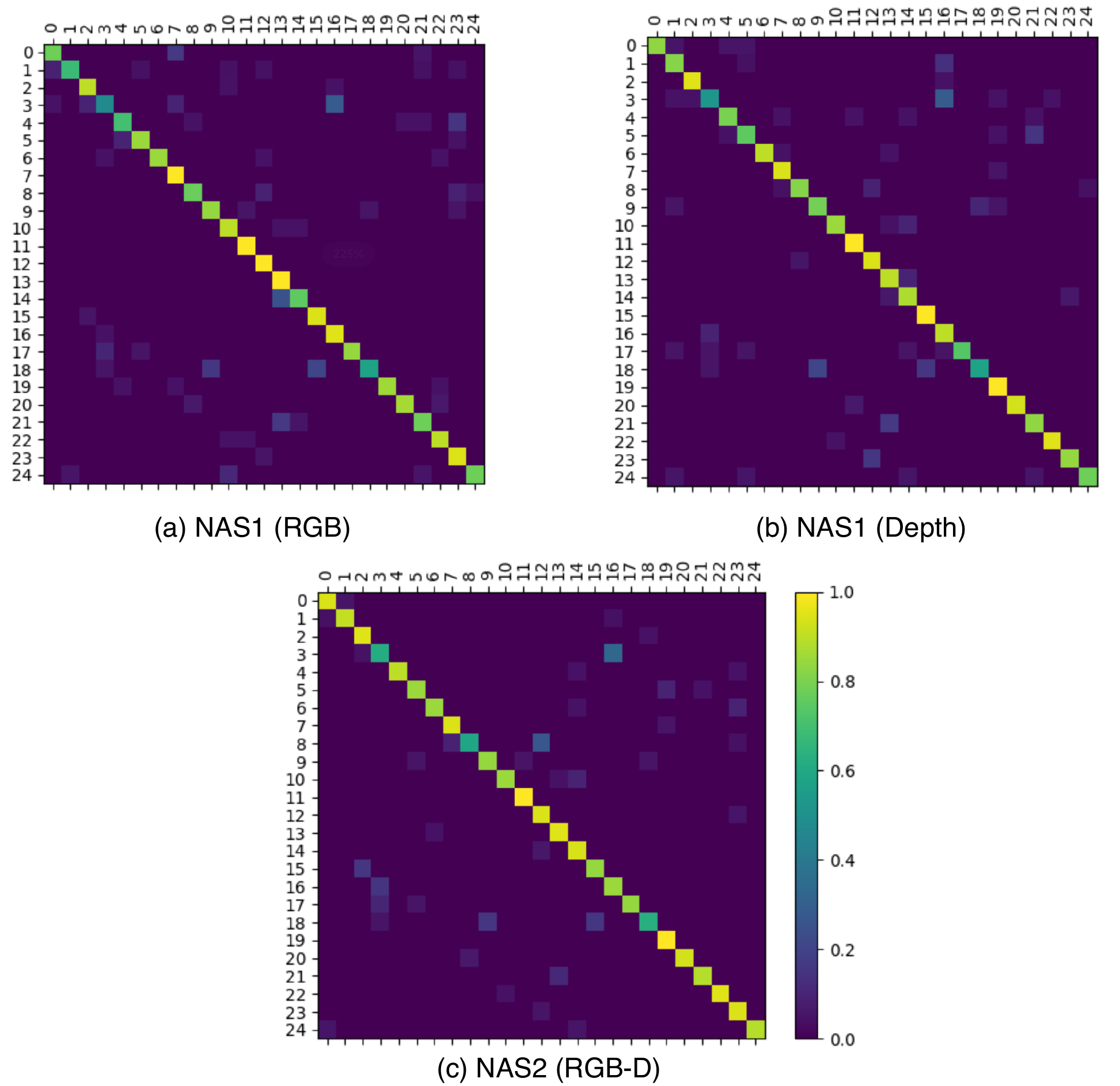}
  \caption{\small{
  The confusion matrices obtained by comparing the ground-truth labels and the predicted labels from the NAS1 and NAS2 networks trained on the NVGesture dataset. Best seen on the computer, in color and zoomed in.}
  }
\label{fig:confusion_matrices}
\end{figure}

Fig.~\ref{fig:confusion_matrices} evaluates the coherence between the predicted labels from the searched `NAS1' and `NAS2' architectures, and the ground truths on the NVGesture dataset. The coherence is calculated by their confusion matrices. We observe that with RGB-D-temporal synergies, `NAS2' has less confusion between the input classes and provides generally a more diagonalized confusion matrix. This improvement is better observed in the first three and last six classes.

\textbf{Results on EgoGesture.} \quad   We also evaluate the robustness of our searched architectures on egocentric (first-person view) gesture recognition. Compared with the the third-person view gesture recognition, the main challenges include the complex scene background and motion blurriness caused by the subject walking. It can be seen in Table~\ref{tab:EgoGesture} that our approach achieves the best performance (`NAS2' with 94.38\% and `NAS1+NAS2' with 95.52\%) on the EgoGesture dataset using RGB-D modalities, which indicates the strong generalization ability of the proposed 3D-CDC and two-stage NAS method. Note that ResNeXt-101 \cite{kopuklu2019real} needs an extra detector to capture the key segments for preprocessing. As our proposed multi-rate and multi-modal network recognizes the gesture on original video clips directly, the performance might be further boosted with the gesture detector.

\begin{table}[t]
\centering
\caption{Results on the EgoGesture \cite{tang2018multi} dataset.}
\resizebox{0.42\textwidth}{!}{
\begin{tabular}{c c c}

 \toprule
 Method & Modality & Accuracy (\%) \\
 \midrule
VGG16+LSTM \cite{simonyan2014two} &RGB & 74.7\\
C3D+LSTM+RSTTM \cite{tran2015learning} &RGB & 89.3\\
ResNeXt-101 \cite{kopuklu2019real}  &RGB & \textbf{93.75}\\
\textbf{NAS1 (Ours)} &RGB & 93.31\\ 
\midrule
VGG16+LSTM \cite{yang2014super}  &Depth &77.7 \\
C3D+LSTM+RSTTM \cite{molchanov2016online} &Depth & 90.6 \\
ResNeXt-101 \cite{kopuklu2019real}  &Depth & 94.03\\
\textbf{NAS1 (Ours)}  &Depth & \textbf{94.13}\\ 
\midrule
VGG16+LSTM \cite{carreira2017quo} &RGB-D & 81.4 \\
C3D+LSTM+RSTTM \cite{abavisani2019improving}  &RGB-D & 92.2 \\
I3D \cite{gupta2019progression}  &RGB-D & 92.78 
\\
MTUT$^{F}$ \cite{gupta2019progression}  &RGB-D & 93.87 
\\
\textbf{NAS2 (Ours)}  &RGB-D &94.38 \\  
\textbf{NAS1+NAS2 (Ours)}  &RGB-D &\textbf{95.52} \\
 \bottomrule
\end{tabular}
}
\label{tab:EgoGesture}
\end{table}

\begin{table}[t]
\centering
\caption{Results on the RGB-D action recognition dataset THU-READ \cite{tang2018multi} with the cross-subject protocol. The architectures of `NAS1' and `NAS2' are searched on IsoGD and then retrained/evaluated on THU-READ.} \label{tab:ResultAction}
\resizebox{0.45\textwidth}{!} {\begin{tabular}{l c c c c} 
 \toprule
 Method & Modality & Accuracy(\%)  & \\
 \midrule
 HOG~\cite{wang2009evaluation} & RGB & 39.93 \\
 HOF~\cite{laptev2008learning} & RGB & 46.27 \\
 Appearance Stream~\cite{simonyan2014very} & RGB & 41.90 \\ 
 C3D~\cite{tran2015learning} & RGB & 66.25\\ 
 SlowFast~\cite{feichtenhofer2019slowfast} & RGB &69.58 \\ 
 \textbf{NAS1 (Ours)} & RGB & \textbf{71.25} \\ 
 \midrule
 HOG~\cite{wang2009evaluation} & Depth & 45.83 \\
 HOF~\cite{laptev2008learning} & Depth & 43.96 \\
 Depth Stream~\cite{simonyan2014very} & Depth & 34.06 \\
 C3D~\cite{tran2015learning} & Depth &63.75 \\ 
 SlowFast~\cite{feichtenhofer2019slowfast} & Depth &68.75 \\ 
 \textbf{NAS1 (Ours)} & Depth & \textbf{69.58} \\ 
 \midrule
 MDNN\cite{tang2018multi} & RGB-D-Flow & 62.92 \\
 C3D~\cite{tran2015learning} & RGB-D & 75.83 \\ 
 SlowFast~\cite{feichtenhofer2019slowfast} & RGB-D & 76.25 \\ 
 
  \textbf{NAS2 (Ours)} & RGB-D &73.85 \\ 
   \textbf{NAS1+NAS2 (Ours)} & RGB-D & \textbf{78.38} \\ 

 \bottomrule
 \end{tabular}}
\end{table}

\section{Discussion and Visualization}
\label{sec:Visualization}
In this section, we first discuss the transferability of the proposed two-stage NAS and 3D-CDC on action recognition task, which is interesting and necessary because it exists task-oriented biases (gesture recognition is less relied on the scene but more related to the fine-grained temporal cues when compared with the action recognition). Then, we analyze the visualization results of the searched architecture and feature response, which are shown in \href{https://github.com/ZitongYu/3DCDC-NAS}{https://github.com/ZitongYu/3DCDC-NAS}.

\subsection{Task Transferability}

\textbf{Generalization to RGB-D Action Recognition.} \quad   In order to validate the generalization ability of our 3D-CDC based two-stage NAS, we transfer the searched architecture (`NAS1' and `NAS2') to another multi-modal video understanding task, i.e., RGB-D action recognition. Here one of the largest RGB-D egocentric dataset THU-READ \cite{tang2018multi} is used for experiments. It consists of 40 different actions and 1920 videos. We adopt the released leave-one-split-out protocol. For fair comparison, C3D ~\cite{tran2015learning}, SlowFast ~\cite{feichtenhofer2019slowfast}, `NAS1', and `NAS2' are pre-trained on Jester gesture dataset. Table~\ref{tab:ResultAction} compares the performance of our method with SOTA methods on THU-READ. It can be seen that our approach outperforms the mainstream 3DCNNs (e.g., C3D~\cite{tran2015learning} and SlowFast~\cite{feichtenhofer2019slowfast}) with a convincing margin, indicating that the architecture searched on the source task (gesture recognition) could be generalized well on the target video understanding task (e.g., action recognition).

\begin{figure}
\centering

\includegraphics[scale=0.65]{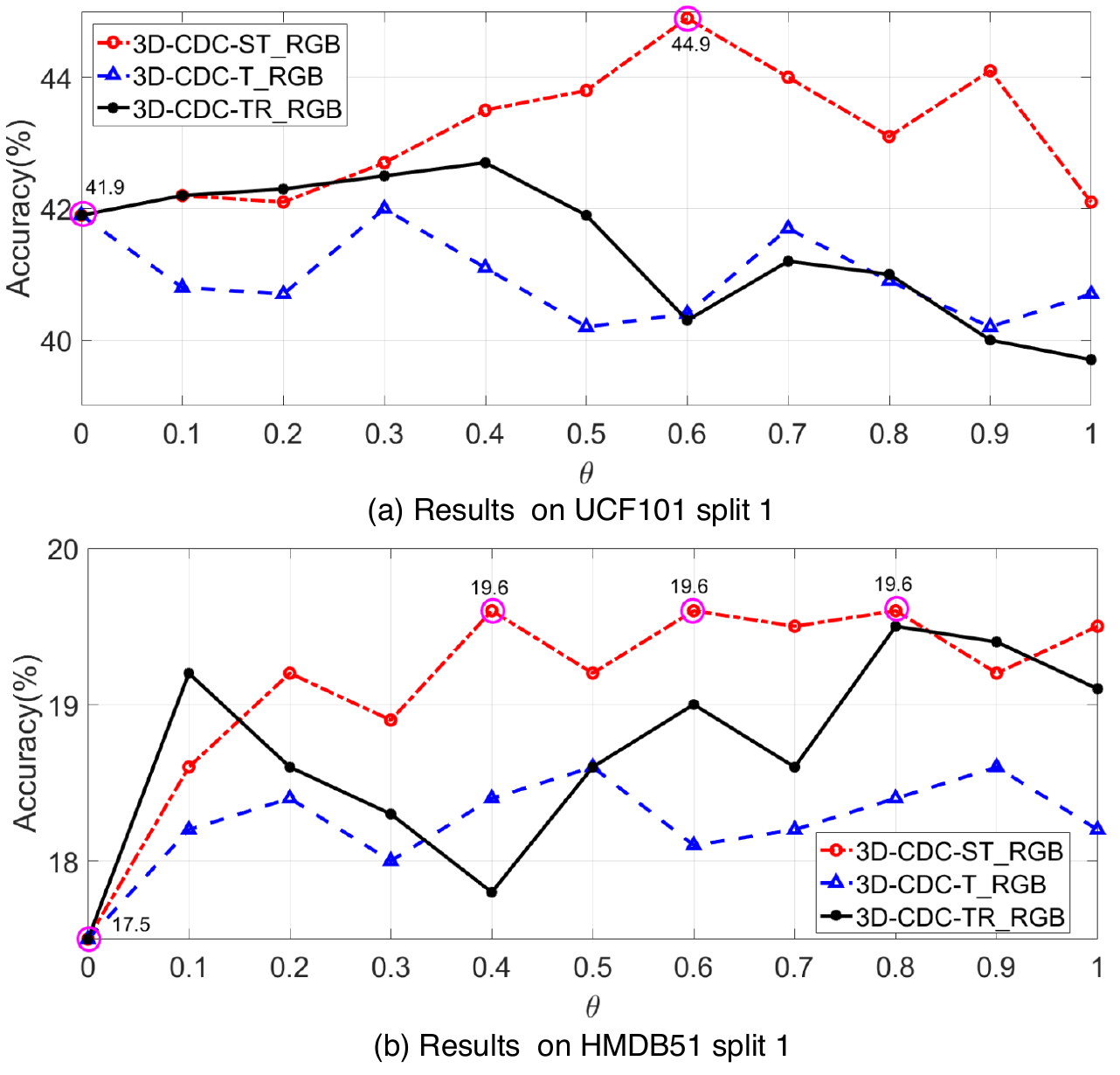}
  \caption{\small{
  Impacts of the 3D-CDC family on two benchmark action recognition datasets (a) UCF101 \cite{soomro2012ucf101}, and (b) HMDB51 \cite{kuehne2011hmdb}. We use 3D-ResNet18 \cite{hara2018can} as the backbone and sequence size 3$\times$16$\times$112$\times$112 as the inputs. All experiments are trained from scratch for fair comparison.}
  }
 
\label{fig:UCF101}
\end{figure}

\textbf{Impact of 3D-CDC for RGB Action Recognition.} \quad  Here we explore the effectiveness of 3D-CDC for scene-based action recognition tasks. Fig.~\ref{fig:UCF101} illustrates the results of two classical scene-related action datasets (UCF101 \cite{soomro2012ucf101} and HMDB51 \cite{kuehne2011hmdb}). It is obvious that compared with the 3D vanilla convolution ($\theta=0$), 3D-CDC-ST improves the performance dramatically in both datasets (+3\% for UCF101 when $\theta=0.6$ and +2.1\% for HMDB51 when $\theta=0.4,0.6,0.8$). The reason might be twofold. On one hand, an enhanced spatio-temporal context is helpful to represent scene-aware appearance and motions. On the other hand, the spatio-temporal difference term can be regarded as a regularization term to alleviate overfitting. Another highlight is that, without extra parameters, 3D-ResNet18 assembled with 3D-CDC-ST outperforms that with Spatio-Temporal Channel Correlation (STC) Block \cite{diba2018spatio} by +2.1\% on UCF101 split 1. In contrast, 3D-CDC-T performs the worst because of its weak spatial context representation capacity and vulnerability to the scene noises, which are vital in these two scene-aware datasets.

\subsection{Visualization of the Searched Architecture}
Here we give a an the searched cells and lateral connections with the proposed two-stage NAS. It can be seen that there are more `CDC-T-06' based operators in all three RGB branches while more `CDC-TR-03' based operators in the depth branches. This consists with the observations in our ablation study of `Impact of 3D-CDC for Modalities' in Section 4.1. It is interesting to find that the lower-level lateral connections are sparser (i.e., with more `Zero' operators) and the high-level lateral connections have more learnable operators (i.e., convolution operators). This might inspire the video understanding community to design more reasonable multi-modal networks in the future.

\subsection{Feature Visualization}
\label{sec:Visualization2}
The neural activation (before Pool3 in C3D) are visualized. It can be seen that the proposed 3D-CDC-ST, 3D-CDC-T, and 3D-CDC-TR  enhance the spatio-temporal representation and enforce the model to focus more on the trajectories of arms and hands. As for the depth modality, all the convolutions are able to make the accurate attention on the movements from arms and hands due to the benefits from the foregrounds provided by the depth inputs. Despite more robust representation achieved by the 3D-CDC family, the interferences from the sensor-based noise and undesirable movements (e.g., head and hair) still occur. Thus, it is necessary to explore RGB-D-temporal synergies to overcome such limitation.

\section{Conclusion}
\label{sec:conclusion}
We present a novel 3D convolution family called 3D-CDC for enhancing the spatio-temporal representation for video understanding tasks. Over the 3D-CDC search space, we propose a two-stage NAS method to discover well-suited multi-rate and multi-modal networks with RGB-D-temporal synergies. Extensive experiments show the effectiveness of our method. Future directions include: 1) exploring 3D-CDC family on other video understanding tasks (e.g., temporal localization); 2) searching temporal synergies with more modalities (e.g., audio and skeleton).

\section{Acknowledgment}
This work was supported by the Academy of Finland for project MiGA (grant 316765), ICT 2023 project (grant 328115), Infotech Oulu, and the Chinese National Natural Science Foundation Projects $\#$61961160704, $\#$61876179, Science and Technology Development Fund of Macau No. 0025/2019/A1. The authors also wish to acknowledge CSC-IT Center for Science, Finland, for computational resources.

\ifCLASSOPTIONcaptionsoff
  \newpage
\fi

\bibliographystyle{IEEEtran}
\bibliography{IEEEabrv,reference}

\end{document}